\pdfoutput=1
\documentclass[11pt]{article}

\usepackage[]{EMNLP2022}

\usepackage{times}
\usepackage{latexsym}

\usepackage[T1]{fontenc}
\usepackage[utf8]{inputenc}

\usepackage{microtype}

\usepackage{inconsolata}

\usepackage{tikz}
\usepackage{booktabs}
\usepackage{multirow}
\usepackage{comment}
\usepackage{subcaption}
\usepackage{amsmath,amssymb}

\title{Describing Sets of Images with Textual-PCA}

 \author{Oded Hupert\qquad{\bf Idan Schwartz}\qquad{\bf  Lior Wolf}
         \\School of Computer Science, Tel Aviv University}
\begin{document}
\maketitle
\begin{abstract}
  We seek to semantically describe a set of images, capturing both the attributes of single images and the variations within the set. 
  Our procedure is analogous to Principle Component Analysis, in which the role of projection vectors is replaced with generated phrases. First, a centroid phrase that has the largest average semantic similarity to the images in the set is generated, where both the computation of the similarity and the generation are based on pretrained vision-language models. Then, the phrase that generates the highest variation among the similarity scores is generated, using the same models. The next phrase maximizes the variance subject to being orthogonal, in the latent space, to the highest-variance phrase, and the process continues. Our experiments show that our method is able to convincingly capture the essence of image sets and describe the individual elements in a semantically meaningful way within the context of the entire set. Our code is available at: \url{https://github.com/OdedH/textual-pca}.
\end{abstract}

\section{Introduction}

Given a set of images with a common theme, it seems to be extremely easy for humans to identify and describe the common theme. While computer algorithms can identify in-set and out-of-set images using anomaly detection methods~\cite{scholkopf1999support,golan2018deep}, describing the common theme seems more challenging.

Captioning methods~\cite{Mao2014DeepCW, li2020oscar, tewel2021zero,li2022blip} are extremely effective in describing single images. However, one cannot directly employ such a method to the mean image representation, in hope of describing a set of images. Since image captioning engines are trained to be specific and not to provide general terms, the resulting captions would not be generic enough. For example, images of people are described by image captioning methods as ``man'', ``woman'', ``child'', etc., and not by generic terms, such as ``person''. To create a representation of an image set, one has to employ higher-level themes. Unable to do so, image captioning methods output non-grammatical phrases, which include, for example, phrases such as ``cat dog'' for sets that contain both pet species.

Our first contribution is to retool the BLIP~\cite{li2022blip} image captioning tool to perform the task of image-set captioning. This is done through modifying the autoregressive process of BLIP without retraining the underlying network. WordNet~\cite{miller1995wordnet} is used to  manipulate the likelihood of words by reducing the likelihood of hyponyms (specific terms) and increasing the likelihood of shared hypernyms (more generic terms). 

\begin{figure*}[t]
	\centering
    \includegraphics[width=1\linewidth]{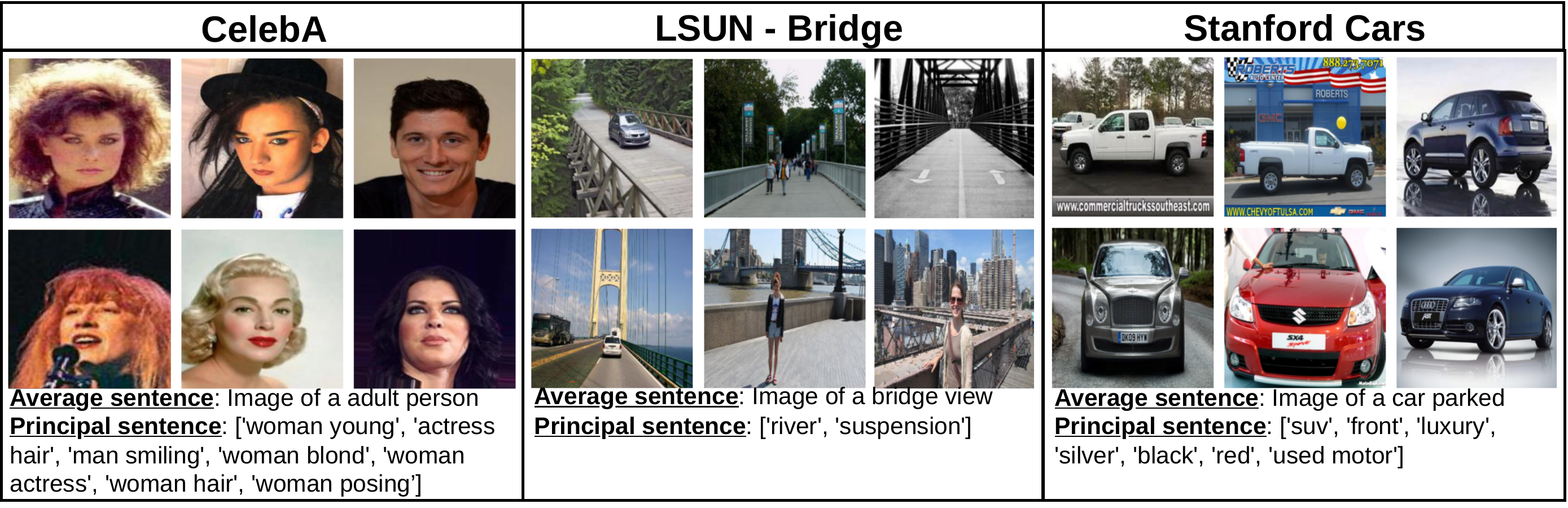}
    \caption{Examples of our textual PCA. For each dataset, we present an average phrase that describes the image set and the principal phrases, i.e., phrases that maximize the variance of the image-to-text matching score subject to being mutually orthogonal in the embedding space.\label{fig:teas}}
\end{figure*}

Once a common theme is generated for the image set, we seek to identify the directions of variation within the set. This way, we can position the set elements in the context of the entire set. For example, staying with the example of images of persons, the images can vary by pose, age, hair style, facial expression, etc.

Motivated by the PCA method, we seek to find the phrase whose visual-language similarity to the images of the set has the highest variance. Then, once again following PCA, we recover a phrase that maximizes the variance among all phrases that are orthogonal to the first phrase in the textual embedding space.

Our second contribution is, therefore, the ability to extract different phrases that capture directions of semantic variability in the image set. This is also done by retooling BLIP. In this case, the likelihood of the next token combines three terms: (i) the likelihood assigned by BLIP, (ii) a term that maximizes the variance of the similarity between the resulting phrase and the images of the input set, and (iii) an orthogonality term that distances the generated phrase from the previously found phrases.

Our textual PCA method seems to be highly suitable for describing sets of images in an intuitive way that combines the general theme with the modes of variation. Consider, for example, Fig.~\ref{fig:teas}. The average phrase clearly depicts that CelebA is a dataset featuring images of adult persons or that LSUN-Bridge contains images of bridges. From the principal phrases we can learn about the traits of the dataset. For example, hair varies considerably in CelebA, and the LSUN-Bridge images differ mainly in the type of the bridge (suspension) or whether it crosses a river. %We can also see that cars from Stanford cars changes by color.

 {We evaluate} our method on multiple datasets, including existing image datasets, such as CelebA, LSUN, and ImageNet, and on sets of images obtained by applying clustering methods to large image collections. We also show that the obtained projections are more informative than the baselines in predicting attributes in CelebA. Finally, in lineup-type experiments, we show that the projections we create provide enough information for users to identify the images out of the set.

\begin{figure*}[t]
	\centering
    \includegraphics[width=0.8\linewidth]{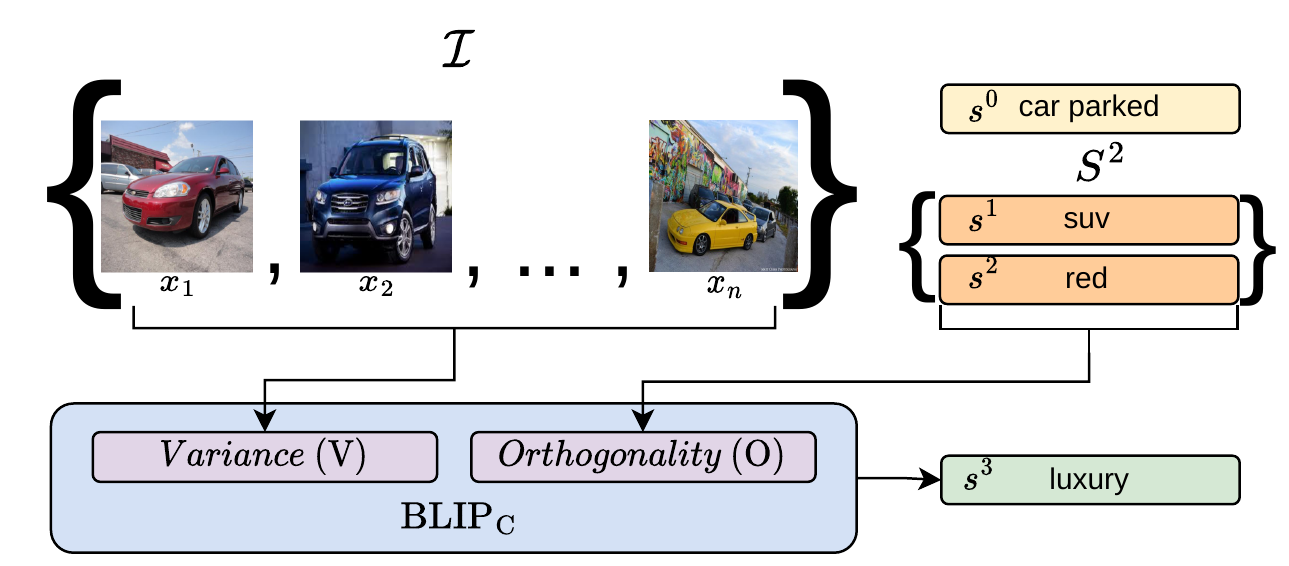}
    \caption{An overview of our approach to generate principal phrases by modifying BLIP's caption head.
    The first step is to create an average phrase, 'car parked' that captures the common features of the set of images $\mathcal{I}$. Then, during the auto-regression process, we consider two types of operators: (i) $\operatorname{V}$ for maximizing the variance over the images, and (ii) $\operatorname{O}$  for generating phrases that are orthogonal to the previous phrases (e.g., `suv', `red'). The result is a novel phrase `luxury' that describes the car's pricing.
    }\label{fig:overview}
\end{figure*}

\section{Related work}

Some of the earliest deep learning attempts in image captioning relied on RNNs with attention~\cite{Mao2014DeepCW,klein2014fisher,xu2015show}, while more recent approaches apply spatial reasoning via graphs~\cite{yao2018exploring,kipf2016semi}, adapt to multi-image processing~\cite{braude2022ordered}, and handle  multi-modal input~\cite{schwartz2019simple}. Annotations by humans are used to train most of the current captioning methods. As human references cannot account for every possible scene, other approaches rely more on web-scale unsupervised image-text datasets~\cite{zhang2021vinvl,devlin2018bert,li2020oscar}. In these approaches, smaller datasets annotated by humans are used as final fine-tuning. Captions based on human annotations can enhance correspondence with human annotators. There is, however, a tendency for them to be repetitive and not particularly informative. 

Recently, several methods that employ web-scale training data directly have been suggested. The first method, ZeroCap, employs CLIP~\cite{radford2021learning}, a prominent web-scale image-text matching model, to guide a pre-trained language model, GPT-2, to caption images~\cite{tewel2021zero, tewel2022zero} without performing any training (``zero-shot''). MAGIC~\cite{su2022language}, is another zero-shot method for image captioning that skews the next-token distribution of a GPT-2 language model to match a given image, based on the CLIP score. Unlike ZeroCap, no gradient updates are applied. BLIP~\cite{li2022blip} applies conventional training (not zero-shot) and jointly learns an image-text metric with a contrastive loss and a caption decoder head.

Also relevant to our work are causality frameworks that study concept discovery, such as TCAV~\cite{kim2018interpretability} and CaCE~\cite{goyal2019explaining}. By intervening on labeled concepts, these approaches study causal effects. However, they require labeled data. Although expertise can be acquired to identify the underlying structure of a problem, the process is still contaminated by human bias. In contrast to these approaches, ours does not require supervision. 

The discovery of visual concepts is also a subject of research. An intuitive approach involves clustering according to segmented regions~\cite{ghorbani2019towards}. A Shapley theorem-inspired method is described in subsequent work~\cite{yeh2019concept}. In a StyleGAN model, latent variables that control the semantic properties of images are disentangled~\cite{stylegan}. Consequently, a disentangled StyleSpace was proposed for finding the attributes that determine classification~\cite{lang2021explaining}. Variational auto-encoders can also be used to reveal concepts used to predict classes~\cite{gat2021latent}. These methods involve finding visual concepts, whereas our approach involves finding textual concepts that describe visual content. This shift is not trivial; in existing works, the meaning of each direction is assigned by human observers, based on manually inspecting samples, and not all directions can be easily described.

\section{Method} \label{sec:method}

Our approach is analogous to Principal Component Analysis (PCA), in which the projection directions are replaced by generated phrases, Fig.~\ref{fig:overview}. PCA identifies the vectors that best fit the data, in terms of Euclidean distance between each data point and its reconstructed vector. It can be equivalently defined as finding directions that maximize the variance of the projected data. The principal vectors are, therefore, directions in which data varies, making them informative. This makes them highly useful for data analyses and dimensionality reduction. These vectors are extracted as mutually orthogonal vectors, in order to capture different directions, and for the projections to span the entire data set.

While PCA can be applied to embedded image vectors, we find that it lacks semantic meaning. In this work, we propose Textual-PCA. Formally, our goal is to create a average phrase $s^0$ and a sequence of principal phrases $S^l = \langle s^1,\dots s^l \rangle$, where $l$ is the number of principal phrases. The phrases are aimed to be a concise set that describes a set of images  $\mathcal{I}=\{x_1,\ldots, x_n\}$, where $n$ is the number of images. The phrase $s^0$ captures common traits in $\mathcal{I}$, and the rest capture different modes of variability. In the following sections we discuss how we find fluent principal phrases.

A principal phrase is created by finding a textual direction that captures semantic variance within an image set. Throughout our process, we employ the multi-modal BLIP model~\cite{li2022blip}, which provides textual and image encoders ($E_T$ and  $E_I$, respectively), an image-text metric ($\operatorname{BLIP}_{\text{M}}$), and a captioning head ($\operatorname{BLIP}_{\text{C}}$). 

Given the set of images $\mathcal{I}$, we start the process with the generation of the average phrase $s^0$. This phrase captures the common attributes of all images in the set. As the first step of generating the average phrase, we average the image representations, $\bar x= \frac{1}{n}\sum_{ x_i\in\mathcal{I}} E_I (x_i)$. We then use $\bar x$ to initialize the auto-regressive BLIP's captioning head,
\begin{align}
    s^0_{t+1} = \operatorname{BLIP}_{\text{C}} (s^0_t, \bar x)\,,
\end{align}
where $s^0_t$ is the average phrase of length $t$.

\subsection{Generating the average phrase}
In the average phrase, we aim for the most generic attributes. For instance, we prefer to increase the potential of the ``church'' token over the more specific ``cathedral'' token, if both of them are top tokens. However, image captioning models are trained on specific captions, and, as a result, tend to be very specific. 

We, therefore, intervene in the BLIP auto-regressive generation process by manipulating the likelihood of each token. Based on the WordNet graph~\cite{miller1995wordnet}, we  aggregate specific terms that have similar meanings into more general terms. 

Our algorithm only considers the top 12 most probable tokens and zero out the rest. This value was determined early in the development process and kept unmodified to obtain all results. Captioning models tend to describe items at a certain level dictated by the training data, i.e., use `cat' and not `mammal' or `Persian cat.' This is probably dictated by the basic level at which items are often perceived~\cite{rosch1976basic}. Since the replacement method relies on tokens in the top-tokens list, the average sentence is more specific (e.g., `cat' and not `mammal'). We then iterate across all candidate tokens in the order of their likelihood.

For each candidate token $t$, we consider, among all other tokens, the set $A_1$ that contains tokens $r$ such that $r$ is an immediate ``is-a'' ancestor of $t$ (a direct hypernym). In this set, we consider the token $r$ with the higher probability, and add the probability assigned to $t$ to the probability associated with $r$, while zeroing the probability of $t$. 

If the set $A_1$ is empty, we consider the set $A_2$ that contains tokens $r$ that are direct hyponyms of $t$. We add the probabilities of all $r\in A_2$ to that of $t$, and zero the probabilities of these hyponyms. 

Finally, if set $A_2$ is also empty, we consider the set $A_3$ that contains all tokens $r$ among the tokens such that $t$ and $r$ have a shared direct hyponym $q$. Among the set $A_3$, we select token $r$ with the highest probability and add token $q$ to the set of candidate tokens with a probability that sums the probabilities of both $t$ and $r$. The probabilities of these two tokens are then zeroed.

Two concrete examples, based on real images are: (i) the token $t$=`sofa'.  $A_1$ and $A_2$ are empty. Two other probable words in $A_3$=\{`chair', `bench'\} share the same hypernym $q$=`seat'. The probability of r=`chair' is higher than 'bench', we update $\text{prob}(q)=\text{prob}(r)+\text{prob}(t)$, $\text{prob}(r)=\text{prob}(t)=0$, where $\text{prob}$ is the likelihood of the given token.  (ii) Another example for the word $t$='salad', $A_1$=\{`dish'\}, which is the immediate hypernym of t. $A_3$=\{`pasta', `soup', `curry'...\}  those words share the same hypernym `dish' and are probable tokens. In this scenario, the probability of the token $t$=`salad' would add to the likelihood of the token $r$=`dish' from $A_1$, i.e., $\text{prob}(r)=\text{prob}(r)+\text{prob}(t)$, and $\text{prob}(t)=0$.

\subsection{Generating the principal phrases}

The next principal phrases, which capture variance, are also generated with the $\operatorname{BLIP}_{\text{C}}$ auto-regressive process. We initialize the captioning head again with $\bar x$. During the auto-regressive process, we modify the next token potentials with two terms: (i) $\operatorname{V}$, which maximizes the variance of the generated phrase with the images, and (ii) $\operatorname{O}$, which is responsible for maximizing orthogonality with the previous phrases (WordNet is not used). 

Let $\hat p_{t, k}^{i}$ be the potential of the $i$-th principle phrase of length $t$, $t-1$ of which were already set, to have token $k$ at position $t$, 
\begin{align}
    %\label{eq:ti}
    \nonumber
    \hat p_{t, k}^{i} \propto \exp ( p_{t, k}^{i} + \lambda_v \operatorname{V}(s_{t,k}^i, \mathcal{I})-\lambda_o \operatorname{O}(s_{t,k}^i, S^{i-1})),
\end{align}
where $p_{t, k}^{i}$ is the original distribution of BLIP's captioning head, and $s_{t,k}^i$ is the $i$-th principle phrase after $t-1$ steps, with token $k$ at location $t$. $\lambda_v, \lambda_o\in\mathbb{R}$ are hyperparameters. We compute $\hat p_{t, k}^{i}$ only for the 1000 most likely tokens (those with the highest $p_{t, k}^{i}$), since the potential of the remaining tokens is usually close to zero.

We define the variance operator $\operatorname{V}$ as the sum of the BLIP's matching scores between the token $s_{t,k}^i$ and an image in the set $x\in\mathcal{I}$ minus the average BLIP's matching score, i.e.,
  \begin{equation}
  \nonumber
     \operatorname{V}\left(s_{t,k}^i, \mathcal{I}\right) = \sum_{x\in \mathcal I} (\operatorname{BLIP}_{\text{M}}(s_{t,k}^i,x)-\mu(s_{t,k}^i))^2,
 \end{equation}
where $\mu(s_{t,k}^i) = \frac{1}{n}\sum_{x\in \mathcal{I}}\operatorname{BLIP}_{\text{M}}(s_{t,k}^i,x)$, and $\operatorname{BLIP}_{\text{M}}$ is a BLIP's matching score. Prior to calculating this matching score, we subtract the embedding of the average phrase $s_0$ from the generated phrase embedding. 

The orthogonality term $\operatorname{O}$ serves to emphasize novel phrases that encapsulate the broad set of factors that define an image set, by encouraging orthogonality between the current phrase potential and all previous phrases %by minimizing their dot-product 
in BLIP's phrase embedding space
  \begin{equation}
  \label{eq:orth}
     \operatorname{O}\left(s_{t,k}^i, \mathcal{S}^{i-1}\right) = \sum_{s^j\in \mathcal{S}^{i-1}} E_T(s_{t,k}^i)^\top E_T(s^j),
 \end{equation}
where $E_T$ is BLIP's textual encoder. This way, the principle components do not repeat the description of the set, despite conditioning the generator on the same mean vector $\bar x$.

\begin{figure*}[t]
\centering
\includegraphics[width=.8\linewidth]{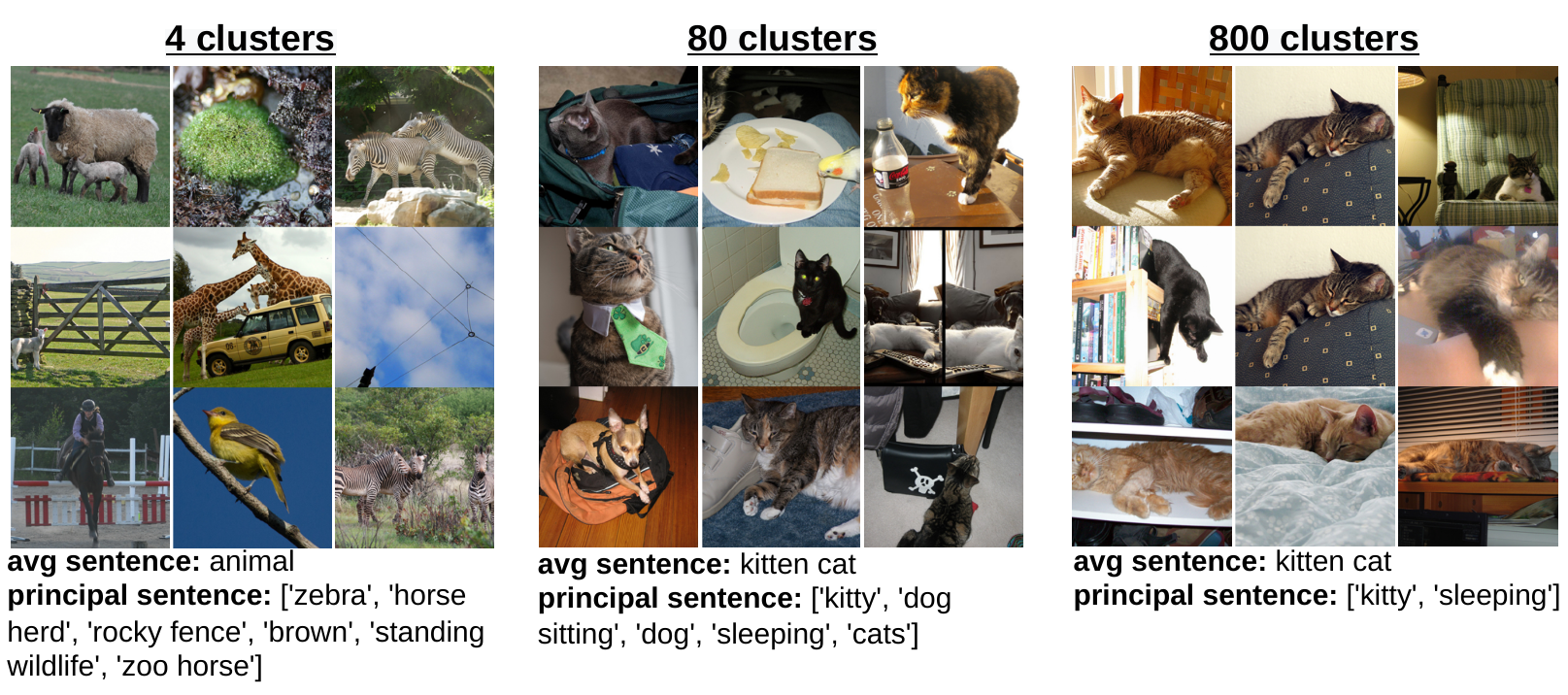}
\caption{Hierarchical clustering of COCO at different granularity levels. As the clusters get smaller, i.e., with more clusters,  they become more homogenous and the principal phrases become  more specific.}
\label{fig:coco}
\end{figure*}

\begin{table*}[t]
\centering
% \resizebox{\linewidth}{!}
%{
\begin{tabular}{lccc} 
\toprule
\textbf{Method} & Named Datasets & COCO & ImageNet\\
\hline
Most Frequent Words & 0.808 $\pm$ 0.14  & 0.835 $\pm$ 0.37 & 0.636 $\pm$ 0.16\\
\hline
ZeroCap+PCA (CLIP space) & 1.068 $\pm$ 0.43  & 0.885 $\pm$ 0.80 & 0.676 $\pm$ 0.30\\
ZeroCap+KMeans (CLIP space) & 1.128 $\pm$ 0.47 & 1.077 $\pm$ 1.49 & 0.586 $\pm$ 0.33\\
MAGIC+PCA (CLIP space) & 1.035 $\pm$ 0.34  & 0.891 $\pm$ 0.78 & 0.666 $\pm$ 0.26\\ 
MAGIC+KMeans (CLIP space) & 1.270 $\pm$ 0.56 & 1.049 $\pm$ 1.14 & 0.685 $\pm$ 0.36\\ 
\hline
ZeroCap+PCA (BLIP space) & 1.290 $\pm$ 0.40 & 1.001 $\pm$ 0.89 & 0.992 $\pm$ 0.33\\
ZeroCap+KMeans (BLIP space) & 1.225 $\pm$  1.13 &  1.225 $\pm$ 1.32 & 0.712 $\pm$ 0.27\\
MAGIC+PCA (BLIP space) & 1.128 $\pm$ 0.19 & 1.004 $\pm$ 0.91 & 0.744 $\pm$ 0.31\\ 
MAGIC+KMeans (BLIP space) & 1.073 $\pm$ 0.4 & 1.051 $\pm$ 1.02 & 0.706 $\pm$ 0.43\\ 
\hline
Ours & \bf 1.515 $\pm$ 0.65 & \bf 1.261 $\pm$ 1.56 & \bf 1.095 $\pm$ 0.47\\
\hline
\end{tabular}
%}
\caption{Variance in the BLIP space, averaged, per image, across all principle phrases. Shown are the mean$\pm$ Standard Deviation in each group of image sets.}
\label{tab:variance}
\end{table*}

\begin{figure*}[t]
	\centering
    \includegraphics[width=1\linewidth]{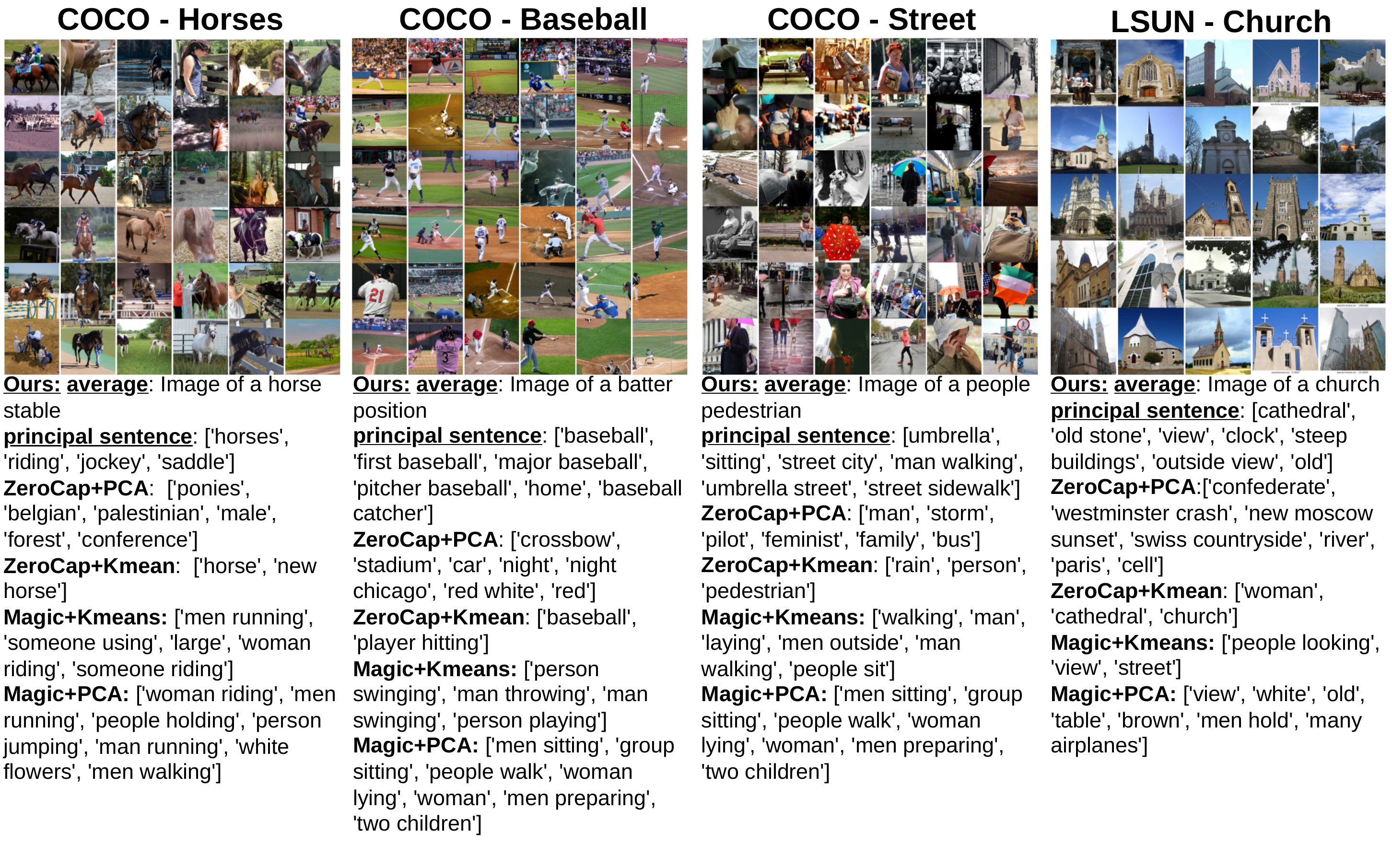}
    \caption{Four image sets: COCO-horses, COCO-baseball, and COCO-Street (names were assigned by us) obtained by clustering COCO, a large visual dataset, and LSUN-Church. For each image set we present the average phrase and principal phrases generated by our method (underlined). For comparison, we present four baselines: ZeroCap+PCA, ZeroCap+KMeans, MAGIC+PCA, and MAGIC+KMeans, obtained in BLIP space.}\label{fig:dataset-short}
\end{figure*}

\begin{figure*}[t]
	\centering
    \includegraphics[width=.8913891\linewidth]{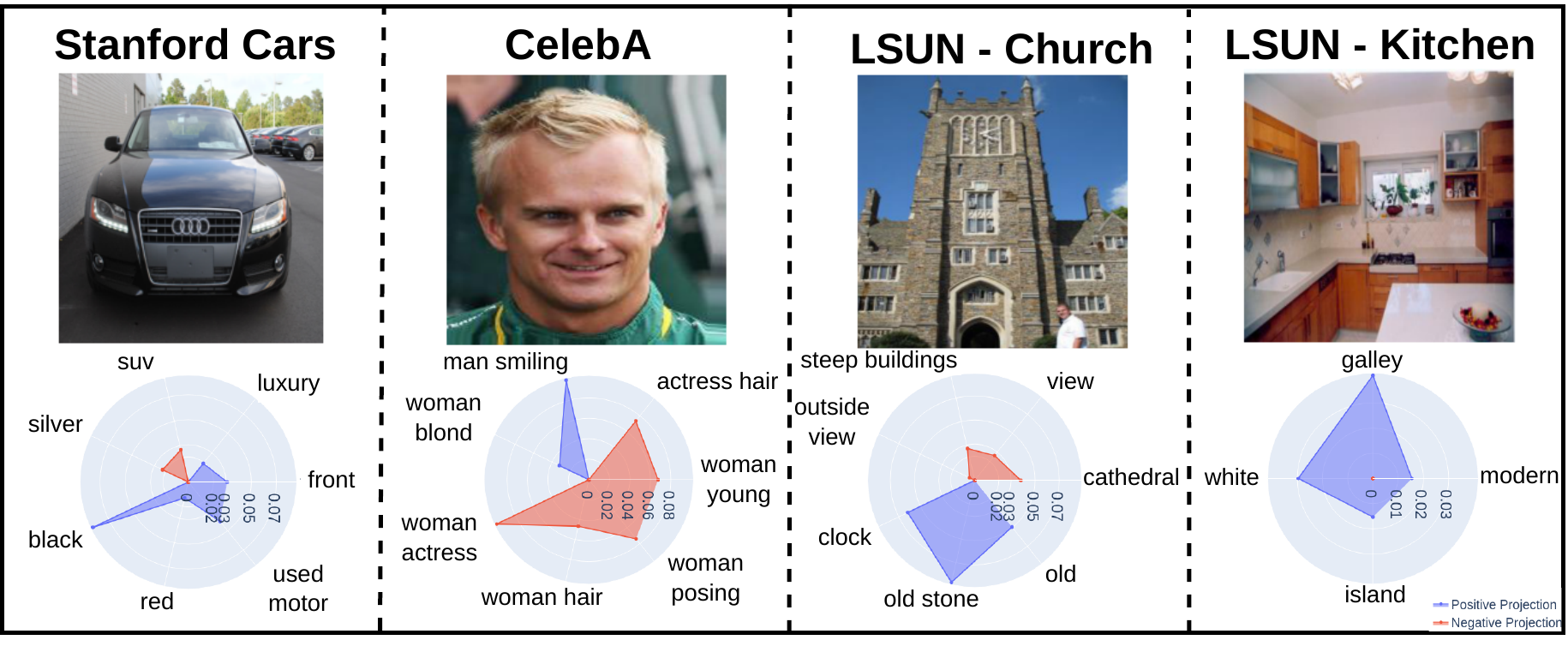}
    \caption{Sample radar plots, in which the value of the projection to the principle phrases are depicted. In blue we show positive correlation and in red negative correlation.}\label{fig:radar}
\end{figure*}

\section{Results}\label{sec:results}

In all experiments, we set the hyperparameter $\lambda_{v}=5$ and $\lambda_{o}=10$. These values were determined early on in the development process and kept unmodified for obtaining all results. 

Evaluation is done on datasets covering a variety of objects and settings (i.e., Named Datasets). CelebA~\cite{celeba} is a large-scale dataset of faces. LSUN~\cite{yu2015lsun} contains ten different scene categories, from which we use the images of bridges, churches, and kitchens. The Stanford cars dataset contains different types of vehicles. 

The first 20 categories of ImageNet~\cite{Deng2009ImageNetAL} were also used. 

We also employ sets that were obtained by hierarchically clustering COCO~\cite{ty2014coco} with an agglomerative algorithm in the CLIP embedding space. 
We cluster until we obtain 80 clusters. This number was selected since there is a small hinge around 80 in the graph depicting the number of clusters vs. the clustering error. Second, with 80 clusters, large enough clusters are starting to form. Fig.~\ref{fig:coco} demonstrates the level of specificity obtained for three levels of clustering. With 4 clusters, one can see a cluster of animals; with 80 clusters, a cluster of cats exists, and with 800 clusters, a subset of the cats with less variation is obtained. Notably, to describe datasets with varying themes, clustering as a first step is a practical technique. 
%For example, we present clusters related to food, horses, baseball, and street view. 

For computational efficiency, from large datasets, such as CelebA or LSUN and some COCO clusters,  we sampled 500 images (each) as the working set.

\noindent{\bf baselines\quad} Our baselines consider sets of vectors in the embedding space of BLIP and then transcribe these into phrases. The sets of vectors are obtained by one of two methods, which are applied to the set of all embedding vectors extracted, using the image encoder, for a given set of images. 

The first method is PCA, which uses Singular Value Decomposition, and the second method is cosine k-means clustering, which was selected since embedding methods have a norm of one. In the case of k-means, the centroids of each cluster are used as the extracted components. 

Note that K-means and PCA are related with relaxation assumptions~\cite{zha2001spectral}. 

%Generating text from CLIP's joint embedding space is not straightforward. We use 
Turning these sets of principle vectors into phrases is done through either the ZeroCap method~\cite{tewel2021zero} or using MAGIC~\cite{su2022language}. Since our method operates in the BLIP encoding space, we created a version of both ZeroCap and MAGIC that are BLIP-based. The results of the unmodified baselines when using CLIP embedding space are also presented. %operates on the BLIP encoding space. 

We also attempted to use BLIP's captioning head to generate text from the principal vectors. However, despite considerable effort, the generation collapsed into the model's default caption. This is most likely due to the sensitivity of the captioning head to distribution changes, i.e., the distribution of PCA vectors differs considerably from the encoding of single images that BLIP was trained on.

We further evaluate a naive baseline of most frequent words. We used BLIP captioning head to generate captions for the images in the set. Then, we created phrases based on the most frequent words.

\begin{table}
\centering
\begin{tabular}{lcc}
\hline
\textbf{Method} & \textbf{CLIP space} & \textbf{{BLIP space}}\\
\hline
ZeroCap+PCA & 0.840 & 0.846\\
ZeroCap+KMeans & 0.813 & 0.823 \\
MAGIC+PCA & 0.849 & 0.840 \\ 
MAGIC+KMeans & 0.852 & 0.829\\ 
\hline
Ours & \bf 0.858 & \bf 0.862\\
\hline
\end{tabular}
\caption{Test accuracy scores for predicting annotated attributes of CelebA dataset, using a simple MLP over the projected values (similarities between the principle phrases and the image).}
\label{tab:celeba}
\end{table}

\noindent{\bf Quantitative Results\quad}

Given a set of principal phrases $S$, and a set of images $\mathcal{I}$, the variability score measures  
\begin{equation}
\label{eq:variance}
|S|^{-1}|\mathcal{I}|^{-1}\sum_{s\in S} \sum_{x\in \mathcal{I}} (E_T(s)^\top E_I(x) - \mu_s)^2,
\end{equation}
where $\mu_s = |\mathcal{I}|^{-1} \sum_{x\in \mathcal{I}} (E_T(s)^\top E_I(x))$. 

The results are presented in Tab.~\ref{tab:variance}. We find that our approach outperforms all baselines.  Note that since PCA in the embedding space maximizes the variance there, any variance lost is a result of translating the principal directions into coherent phrases and back into BLIP space vectors. 

In addition to measuring variance, we also ask to test to what extent principal phrases capture natural traits of the dataset. %We will prefer sentences that maximize variance by conserving information that has semantic meaning over information that has none. 
To this aim, we use the CelebA dataset, a dataset of portrait images of celebrities, since attribute labels for it are available. In CelebA, 40 facial traits are annotated (e.g. hair color, eyeglasses). These are attributes containing semantic information that is not explicitly exposed at the phrase generation stage. 

We consider the generated principal phrases $S_m$ for each method $m$, and embed those phrases back into BLIP, i.e., for each  $s\in S_m$, we compute $E_T(s)$. Similarly to PCA, we project each image $x$ of the image set $\mathcal{I}$ by computing  $E_T(s)^{\top}E_I(x)$. Aggregating over all $s\in S_m$, we represent each image $x$ as one vector for each method $m$. Using basic MLPs with one hidden layer, we then predict the CelebA datasets attributes from these vectors. 

Prediction accuracy on the CelebA test set is reported in Tab.~\ref{tab:celeba}. Evidently, our approach achieves a higher test accuracy score than all baselines.

\noindent{\bf Qualitative results\quad}  Sample results for our method can be found in Fig.~\ref{fig:teas},\ref{fig:coco}. The baselines are not shown, but they are not competitive. Fig.~\ref{fig:dataset-short} provides our results for additional datasets, as well as those of the baseline methods.  Comparing the most frequent words baseline on the Cars dataset, our method (variance score of $0.963$)  extracts: 'suv', 'front', 'luxury', 'silver', 'black', 'red', 'used motor'. Most frequent words (variance score of 0.643) extracts: 'parked', 'car', 'front', 'lot', 'parking', 'black', 'red'.  %UPDATE THIS TO EXACT  that our method produces principal sentences when describing specifically collected, homogeneous, datasets. In this part we wish to show that our method also works on a more natural and noisy sets such as COCO. In addition, we also show how the baselines deal with more natural datasets as LSUN-Church. 
Shown are the results on COCO clusters, which are less homogenous than human-created datasets, as well as on the LSUN-Church dataset. Evidently, the average phrase and the principal phrases produced by our method are related to the theme of the dataset. %This is not the case for all baselines, 

%The baseline ZeroCap+KMeans seems to describe the image set very generally for example, `horse' for COCO-horses or `Baseball' for COCO-Baseball. In contrast, 

Our method extracts multiple relevant attributes. It characterized Churches by structural attributes, such as having a clock or steeple, by their age, or by purpose (the cathedrals attribute). Interestingly, in the horse-related cluster, there is a quantity-related term, horses. In COCO-Baseball the phrases refer to the role of the person (`baseball pitcher', `baseball catcher') the league (`major baseball') and the location (`first baseball' or `home'). 

The baselines ZeroCap+KMeans and MAGIC+KMeans give a very general description that is similar to an average phrase, while ZeroCap+PCA and MAGIC+PCA return phrases that are not related to the image set. For example, in COCO-horses, ZeroCap+PCA produces `conference' as a principal phrase and MAGIC+PCA `white flowers'. 

For COCO-Street, all methods identify that many images present a rainy scene and most relate to `street' or 'pedestrian'. The baselines give a more general description of the set, while our method extracts the specific modes of variation. For example, instead of the general term `rainy', we note if an `umbrella' is present in the photo.

\begin{table}[t]
\begin{tabular}{@{}lccc@{}}
\toprule
 & Answered & \multicolumn{2}{c}{Of those answered}\\
 \cmidrule{3-4}
Method            &  & Correct & Incorrect \\
            \midrule
ZeroCap+PCA & 0.25           & 0.33  & 0.67             \\
MAGIC+PCA   & 0.63           & 0.07    & 0.93            \\
Ours        & {\bf 0.98}          &   {\bf 0.83}   & {\bf 0.17}        \\
\bottomrule
\end{tabular}
\caption{Lineup results for matching radar plots with images. Users could decide not to make a choice. The 1st numerical column contains the ratio of questions that the users chose to answer. The other columns depict the success rate out of the answered queries.}
\label{tab:userstudy}
\end{table}

\begin{figure}[t]
\frame{\includegraphics[width=1\linewidth]{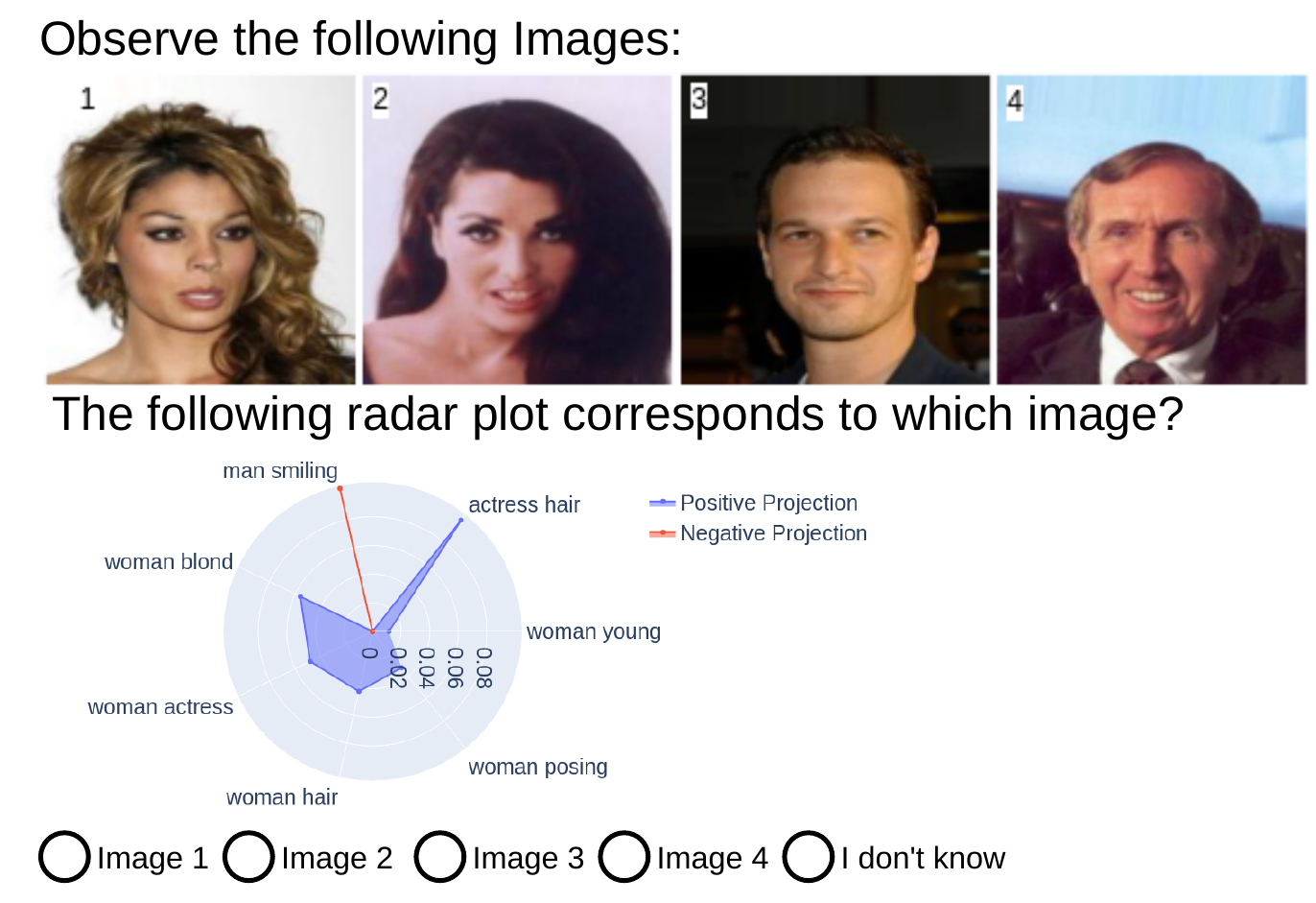}}
\caption{An example query from the lineup user study.}
\label{fig:userstudy}
\end{figure}

%\subsection{User study} \label{subuserstudy}
\noindent{\bf User study\quad} To evaluate the usefulness of the principal phrases in describing images in the context of their image set, we conduct a user study in the form of a lineup. For this, we employ a novel radar plot, which marks both positive and negative projections onto the principal phrases, see Fig.~\ref{fig:radar}.

Each user is given a series of radar plots created by the various methods. For each radar plot, the user is asked to select the matching image out of four options, or to check the option ``I cannot tell'', see Fig.~\ref{fig:userstudy}. The $n=20$ users were first trained using sample radar plots created manually. %The order of presentation is randomized between the methods. 

We compare our method to the two strongest baselines ZeroCap+PCA and MAGIC+PCA, both in BLIP space. The results, listed in Tab.\ref{tab:userstudy}, show that in almost all cases, the users were willing to identify the matching images for our method, while for other methods, they were more reluctant to do so. Out of the choices made, users were able to select the correct image in far more cases for our method than for the baselines.

\noindent{\bf Ablation and parameter sensitivity\quad} %To study the contribution of the different components of our method, an ablation study for our method is conducted. %We explore the effect of modifying our method in different ways.
When generating the average phrase, we aim to generate a general description of the dataset. In order to do so, we use WordNet to aggregate terms to a more general term, as explained~\ref{sec:method}. Without WordNet, the method produces specific phrases that tend to contain recurrent terms, which are less preferable when describing a dataset, as can be seen in Tab.~\ref{tab:ablation_wordnet}.

\begin{table}
\centering
%\begin{small}
\begin{tabular}{@{}lcc@{}}
\toprule
\textbf{Dataset} & \textbf{w/o WordNet} & \textbf{{w/ WordNet}}\\
\midrule
CelebA & woman man & adult person\\
Stanford Cars & car suv  & car parked \\
COCO-Horses & horse horses & horse stable \\ 
LSUN-Church & church cathedral & church\\ 
\bottomrule
\end{tabular}
%\end{small}
\caption{Average phrases when using or not using Wordnet for aggregating terms. We omit the prompt ``Image of a''.}
\label{tab:ablation_wordnet}
\end{table}

\begin{figure}[t]
\centering
\noindent\begin{tabular}{@{}cc@{}}
\includegraphics[width=.481\linewidth]{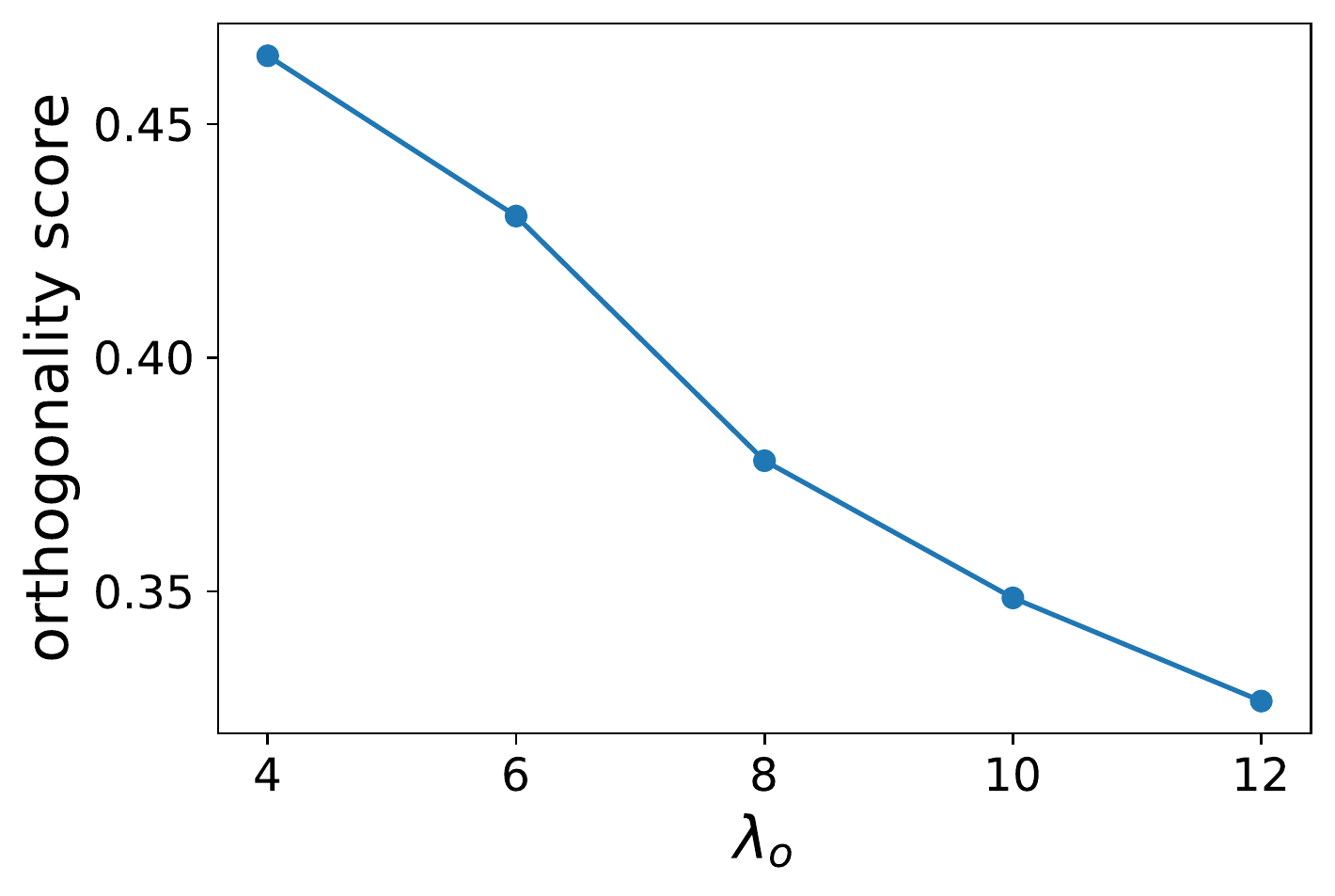}&%\\%(a)\\
\includegraphics[width=.481\linewidth]{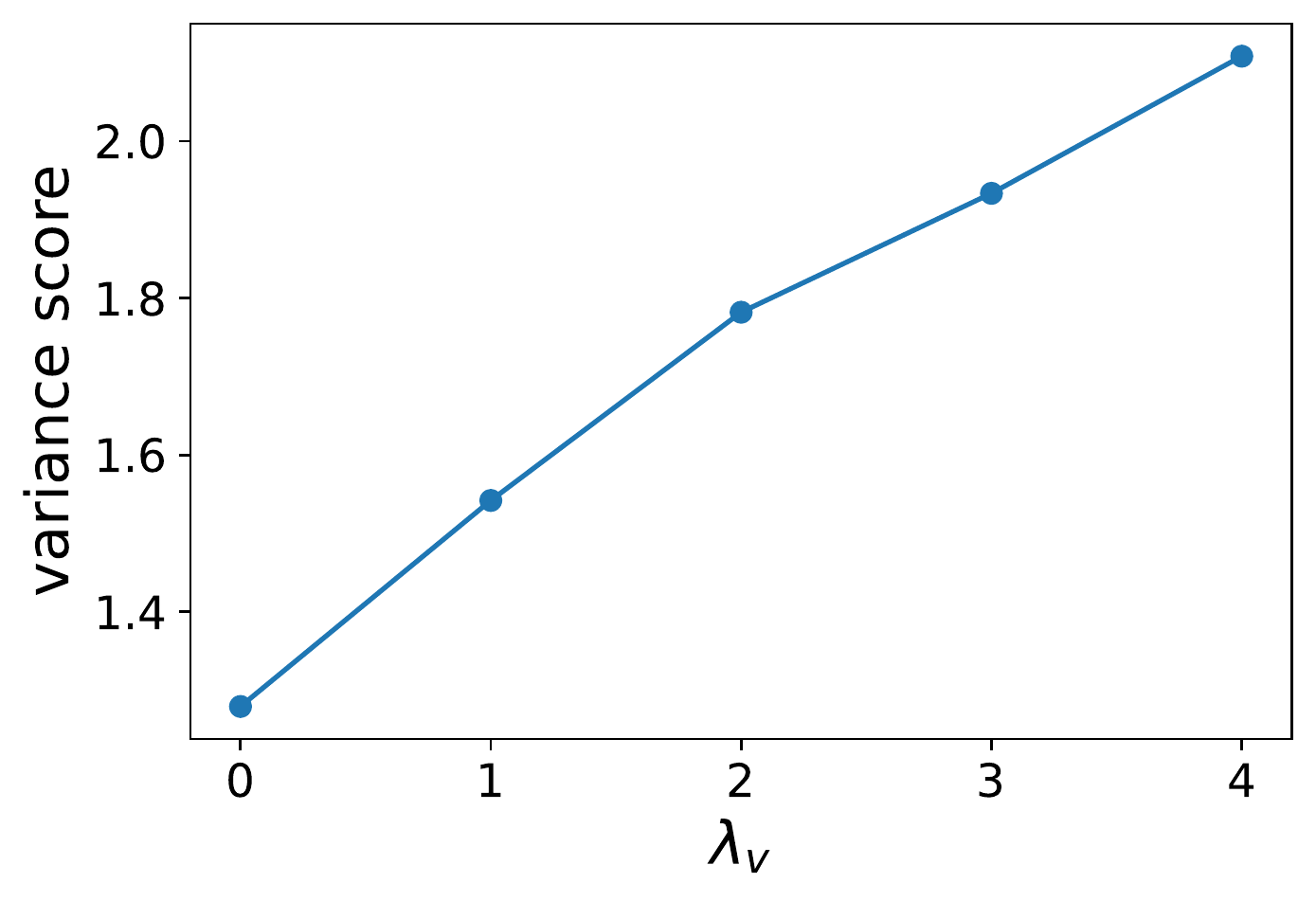}\\%(b)\\
(a) & (b)\\
\end{tabular}
\caption{Hyperparameter sensitivity. (a) The effect of the hyperparameter $\lambda_{o}$ on the mean orthogonality score of the image sets. (b) The effect of the hyperparameter $\lambda_{v}$ on the mean variance score of the image sets.}
\label{fig:ablation_orthogonality}
\end{figure}

The orthogonality coefficient controls the amount of orthogonality between the principal phrases. If we set $\lambda_o=0$, the first principal phrase, which maximizes variance, will be repeated over and over, as we empirically confirmed (there is stochasticity in the method).

If we set it to a very large number, for example $\lambda_o=1000$, we still receive our original first principal phrase, since the orthogonality of one phrase to itself is 0. However, the following phrases lose all connection to the image set in favor of being orthogonal. For example, for LSUN church we will get: ['cathedral', 'johns like', 'overlook sculptures', 'floppy flags', 'daytime narrow', 'sofia framed', 'gravel']. For Stanford cars, instead of ['suv', 'front', 'luxury', 'silver', 'black', 'red', 'used motor'] we get ['suv', 'old monroe', 'frankfurt', 'kidney seller', 'asphalt', 'poles purple'].

In Fig.~\ref{fig:ablation_orthogonality}(a) we quantify how changing $\lambda_o$ effects the orthogonality score (Eq.~\ref{eq:orth}). This is shown for the mean score over all datasets discussed in Sec.~\ref{sec:results} (Named Datasets, COCO, ImageNet). In these experiments $\lambda_v$ is fixed at the default value, and $\lambda_o$ varies from its default value of $10$. As can be seen, for a wide range of $\lambda_o$ values, the orthogonality score is relatively stable.

%\noindent{\bf Variance coefficient\quad}
%Our goal is generating principal sentences that maximizes variance over the set. For that end, we tune a parameter $\lambda_{v}$ as described in Eq.~\ref{eq:ti}. 
The coefficient $\lambda_v$ controls the emphasis on maximizing the variance. If we set $\lambda_{v}=0$ we are left with two other constraints: describing the image set, which is done with BLIP's caption head, and an orthogonality constraint. Therefore, in this case, we obtain meaningful attributes that produce somewhat less variance. As an example, for Stanford Cars, instead of the attributes that maximize and are sorted by variance, ['suv', 'front', 'luxury', 'silver', 'black', 'red', 'used motor'] we get ['silver', 'front', 'luxury used']. Similarly, for CelebA, instead of ['woman young', 'actress hair', 'man smiling', 'woman blond', 'woman actress', 'woman hair', 'woman posing’], we get ['hair hairs', 'young premiere', 'man woman', 'press', 'hollywood photo', 'young', 'man']. If we set $\lambda_{v}$ to a very large number, $\lambda_{v}=1000$, it will produce the same principal phrase that maximizes variance, since it will overcome orthogonality. For Stanford Cars we now get ['suv truck'] and for CelebA ['female'].

Fig.~\ref{fig:ablation_orthogonality}(b) shows how changing $\lambda_{v}$ effects the variance score calculated by Eq.~\ref{eq:variance}. The results are shown for the mean score over all datasets discussed in~\ref{sec:results} (Named Datasets, COCO, ImageNet). Despite varying the coefficient over a wide range (the default value we use is 5), the variance remains in a narrow band that outperforms the baseline methods. This includes the case of $\lambda_{v}=0$ discussed above. We, therefore, conclude that having a related text and orthogonal phrases already lead to the desired PCA effect. Adding the variance maximization term further improves results.

\section{Discussion and limitations}
\label{sec:discussion}
We identify a few ways in which our results could be extended. %These suggestions, however, are left for future work.
First, similarly to other generative tasks, evaluation of the results is not straightforward. Since describing an image set in a way that encompasses both the common theme and the modes of variation is a novel task, there is no established methodology for this evaluation. We believe that the lineup experiments presented demonstrate that humans are able to relate the obtained projections to the images. Given more resources to train human annotators, it would be interesting to obtain human phrases for both the theme and the variation and evaluate the degree to which these would match the method's results.  
Second, our method could help create more informative image captions by including the information of the radar plots we present within the generated text. This way, a rich and descriptive captioning, which addresses the common variations from in-set images with a common theme, would be created. 

Finally, there is no reason not to apply our method to sets of phrases or paragraphs, by replacing CLIP with a summarization engine, such as those based on transformers~\cite{vaswani2017attention}. Distancing ourselves even further from the current work, we note that with the advent of powerful image generation engines~\cite{ramesh2021zero} the role that images and text play in our work could be reversed. Images can visually capture the common theme of a set of phrases or paragraphs as well as their modes of variation.

\section{Conclusions}

Dimensionality reduction methods capture the most significant information of the input vectors using a smaller set of variables. However, these modes are free from semantic constraints and often mix multiple attributes, due to correlations that exist in the data. In this work, we follow in the footsteps of PCA, perhaps the most widely used dimensionality reduction method, and propose a method for extracting orthogonal semantic directions that describe a set of images in the latent space of a vision-language model. First, the ``centroid'' phrase, which describes the main theme of the set is extracted. Then, the directions with the highest variability in the vision-language similarity to the images of the set are extracted.

Our solution combines the BLIP image captioning model with information derived from the WordNet graph. An extensive set of experiments demonstrates that the obtained list of semantically-orthogonal phrases accurately describes the set of images given as input.

\section*{Acknowledgments}
 This project has received funding from the European Research Council (ERC) under the European Union's Horizon 2020 research and innovation programme (grant ERC CoG 725974). 

\bibliographystyle{acl_natbib}
\bibliography{anthology,custom}

\appendix

\label{sec:appendix}

\end{document}